\documentclass[12pt]{article}

\usepackage[utf8]{inputenc}
\usepackage[T1]{fontenc}
\usepackage{lmodern}

\usepackage{geometry}
\usepackage{graphicx}
\usepackage{booktabs}
\usepackage{makecell}
\usepackage{amsmath, amssymb}
\usepackage{algorithm}
\usepackage{algpseudocode}

\usepackage{hyperref}
\usepackage{cleveref}

\title{\textbf{FSD-VLN: Fast-Slow Dual-System Modeling for Aerial Long-Horizon Vision-Language Navigation}}
\author{
Xueke Zhu$^{*1}$, Qingyan Meng$^{*1}$, Liutao Yu$^{1}$, \\
Wei Zhang$^{1}$, Zhengyu Ma$^{1}$, Huihui Zhou$^{1,2}$, Yonghong Tian$^{1,3}$\\[0.3em]
$^*$ These authors contributed equally to this work. \\
$^1$ Pengcheng Laboratory \\
$^2$ Shenzhen Institutes of Advanced Technology \\
$^3$ Peking University
}
\date{}

\begin{document}



\maketitle

\begin{abstract}
  Vision-Language Navigation (VLN) enables unmanned aerial vehicles (UAVs) to navigate autonomously in unfamiliar environments by grounding natural language instructions in real-time vision observations. Compared to traditional GPS-based or pre-programmed navigation methods, VLN offers more intuitive human-machine interaction and greater adaptability. Effective UAV-VLN systems typically require the integration of high-level semantic reasoning and low-latency flight control.
  However, existing approaches often suffer from a structural inconsistency between global multimodal understanding and temporally coherent action generation, leading to unstable trajectories and significant decision latency during long-horizon navigation.
  To address this challenge, we propose FSD-VLN, an efficient fast-slow dual-system framework that explicitly decouples high-level semantic reasoning from low-latency action generation. 
  The framework consists of two asynchronous pathways: a slow system that extracts stable semantic priors from a pretrained vision–language model, and a fast system built upon a Diffusion Transformer (DiT) that models action distributions for flight command generation. This decoupled design enables stable multi-step prediction and improves trajectory consistency by explicitly capturing cross-temporal action dependencies.
  Furthermore, a time-aware adaptive optimization strategy is designed to enhance long-horizon training stability and mitigate gradient oscillations during optimization.
  Extensive experiments in large-scale simulated low-altitude environments demonstrate that FSD-VLN achieves up to $2\times$ higher success rate for navigation in unseen environments compared to prior arts, while reducing both single-action inference latency and overall task execution time by over $50\%$. 
  These results highlight the importance of explicitly modeling the cooperation between semantic reasoning and temporally coherent control,  providing a principled solution for long-horizon aerial VLN.
\end{abstract}

\begin{figure}[tb]
  \centering
  \includegraphics[width=\linewidth,page=1]{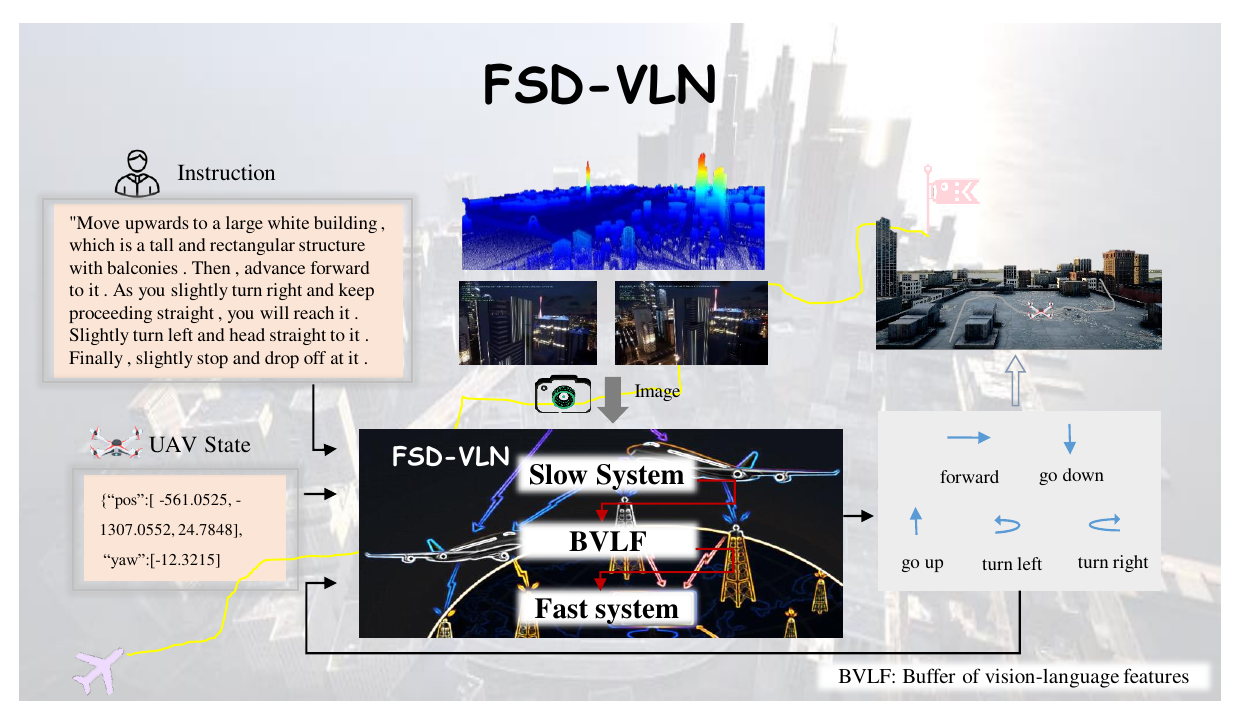}
  \caption{The navigation pipeline of the proposed FSD-VLN framework. During navigation, the slow system utilizes a pre-trained vision-language model to encode visual observations and language instructions into latent embeddings. The fast system adopts a variant of Diffusion Transformer (DiT), consisting of alternating cross-attention and self-attention modules, to process the UAV state embedding sequence and the visual-language embeddings produced by the slow system, respectively. The fast-slow dual system operates asynchronously to generate low-latency UAV actions.
  }
  \label{fig:FDM-VLM}
\end{figure}

\section{Introduction}
Unmanned Aerial Vehicles (UAVs) are being extensively utilized in a broad spectrum of missions, from environmental monitoring to emergency response, requiring robust decision-making in different scenarios.
Traditional UAV navigation often relies heavily on maps or manually designed control strategies\cite{orvsulic2021flying, ubaid2023uavs, blochliger2018topomap, mur2017orb, hornung2013octomap, yin2017improved, janis2016path}, making it difficult to adapt to complex, dynamic, and semantically rich task requirements in open environments.
Recently, the remarkable breakthroughs in large-scale foundation models have catalyzed a paradigm shift in robot intelligence, enabling agents to perform sophisticated multi-modal reasoning. In particular,
vision-language navigation (VLN) enables embodied agents to execute natural language instructions through visual perception and sequential decision-making\cite{hong2021vln, chen2021history}. This paradigm not only enhances the flexibility of UAVs in interactive tasks but also provides a more intuitive and scalable solution for achieving truly intelligent aerial autonomy.

Despite the proficiency of large multimodal models in semantic alignment and zero-shot instruction following \cite{chen2025chatfly,zhang2025grounded, liu2023aerialvln,irshad2021hierarchical}, long-horizon aerial VLN still faces fundamental challenges.
In such navigation tasks, UAVs are required to continuously reconcile high-level semantic reasoning with high-frequency motion control\cite{gao2025openfly, lin2025openvln,zhang2025citynavagent,zhao2025aerial}.
This requirement introduces a structural conflict: whereas global reasoning benefits from computationally intensive multi-modal perception with large-scale models, maintaining flight stability necessitates low-latency non-blocking decision making.
Existing methods struggle to adequately balance the two competing objectives.
Most current VLN methods either adopt reactive action prediction based on instantaneous vision-language features\cite{qi2020object, lin2022adapt,krantz2023iterative}, or use large-scale autoregressive inference models that directly output control commands \cite{zitkovich2023rt,kim2024openvla,black2024pi0,kawaharazuka2025vision}.
Reactive models are vulnerable to short-term perceptual biases, leading to unstable trajectories in long-horizon navigation.
In contrast, large-scale inference models improve semantic consistency but introduce significant inference latency, making them difficult to apply for real-time aerial deployment\cite{gao2025openfly,chen2019touchdown,lee2024citynav,zhou2024navgpt2,zhou2024navgpt}.
Furthermore, traditional action predictors typically process each timestep independently, ignoring the temporal action dependencies that are critical for maintaining flight stability and mitigating long-horizon error accumulation\cite{gao2025openfly,xiao2025uav}.

In this work, to circumvent the reasoning-latency trade-off, we propose FSD-VLN (\cref{fig:FDM-VLM}), an efficient fast-slow dual-system framework that explicitly decouples global semantic understanding from low-level action generation.
The slow system focuses on extracting stable semantic priors from visual-language inputs, while the fast system asynchronously generates flight actions that maintain inherent cross-temporal dependencies.
This design enables UAVs to maintain global navigation consistency while producing low-latency and smooth control signals.
Specifically, we introduce a Diffusion Transformer (DiT)-based action generator that models temporally structured action distributions conditioned on both semantic priors and historical actions.
By explicitly incorporating cross-temporal action priors, the model mitigates distribution drift and enhances trajectory continuity during long-distance navigation.
Furthermore, we propose a time-weighted optimization strategy to stabilize long-horizon action learning, alleviating gradient oscillation and improving convergence robustness.
Extensive experiments conducted in large-scale simulated low-altitude environments demonstrate that FSD-VLN significantly improves navigation accuracy and trajectory quality in both seen and unseen scenarios.
More importantly, our framework substantially reduces single-step decision latency and overall task execution time compared to prior works, achieving a favorable balance between semantic reasoning capability and real-time control efficiency.
Our contributions include:
\begin{itemize}
    \item[1.] We propose an asynchronous dual-system architecture for UAV navigation. By decoupling high-level semantic reasoning from reactive execution, our framework concurrently achieves deep environmental perception and low-latency responsiveness.
    \item[2.] We design a time-weighted optimization strategy that balances the learning objectives across different temporal horizons. By assigning adaptive weights to sequential actions, this strategy effectively mitigates gradient oscillations and enhances convergence robustness. 
    \item[3.] Through intensive benchmarking in diverse low-altitude environments, we show that FSD-VLN outperforms prior arts with a 200\% improvement in navigation success rate. At the same time, it cuts task duration by half, providing a principled and efficient solution for deploying VLN models on agile aerial platforms.
\end{itemize}

\section{Related Work}

\subsection{Vision-Language Navigation for Aerial Platforms}
VLN aims to enable embodied agents to execute natural language instructions through visual perception and sequential decision-making.
With the rapid advancement of Large Language Models (LLMs)\cite{radford2018improving,radford2019language,brown2020language,achiam2023gpt,bai2023qwen,chu2024qwen2,touvron2023llama,zhang2023llama} and Vision-Language-Action (VLA) frameworks 
\cite{chen2019touchdown,zitkovich2023rt,kim2024openvla,huang2023voxposer,anderson2018vision,fried2018speaker,tan2019learning}, multimodal semantic alignment capabilities have improved substantially, further stimulating research on UAV-based VLN.

Early aerial VLN work, such as AerialVLN\cite{liu2023aerialvln}, established the task paradigm of language-guided flight, providing foundational benchmarks for UAV navigation.
Subsequent approaches incorporated LLMs to enhance reasoning ability.
For instance, NavGPT\cite{zhou2024navgpt2} and its extension\cite{zhou2024navgpt} introduced explicit reasoning chains and structured prompting mechanisms to bridge high-level linguistic reasoning with low-level embodied control.
Compared with implicit neural policies, these methods demonstrate stronger interpretability and improved long-horizon planning capability.
More recently, works including OpenFly\cite{gao2025openfly} and VLA-AN\cite{wu2025vla} have improved real-world applicability through key-frame perception modules, staged training strategies, and large-scale simulation environments.
Meanwhile, datasets and simulation platforms such as AVDN\cite{fan2023aerial}, CityNav\cite{lee2024citynav}, OpenFly\cite{gao2025openfly}, and OpenUAV\cite{wang2024towards} have significantly enriched multimodal aerial navigation resources, enabling more complex city-scale evaluations.
Despite these advances, most existing UAV-VLN methods rely on step-wise reactive action prediction or direct autoregressive decoding of control commands.
Such designs often lack explicit modeling of cross-temporal action dependencies, which can lead to trajectory instability and error accumulation in long-distance aerial navigation scenarios.

\subsection{Dual-System Architectures in Autonomous Systems}
Inspired by the dual-process theory, the ``System 1, System 2'' dual-system architectures have been increasingly explored in autonomous driving.
In such framework, system 2 typically leverages large language models to perform high-level reasoning and scene evaluation, while system 1 handles real-time perception and control.
Methods such as LeapAD\cite{mei2024continuously}, ETA\cite{hamdan2025eta}, FASIONAD\cite{qian2024fasionad}, and AdaDrive\cite{zhang2025adadrive} improve adaptability and computational efficiency through asynchronous coordination between high-level planning and low-level execution. 

In embodied robotics, generative control systems including GR00T N1\cite{bjorck2025gr00t} and AgiBot\cite{bu2025agibot} adopt hierarchical or mixture-of-experts structures that explicitly decouple latent planning from continuous control decoding.
More recent work explores temporal frequency decoupling and hierarchical Transformer designs.
For example, HiRT\cite{zhang2024hirt} and Video Prediction Policy\cite{hu2024video} propose a “hierarchical dual-system with high-low frequency collaboration” framework, reducing inference latency while preserving reasoning capability.
In the field of Vision-Language Navigation (VLN), APEX\cite{zhang2026apex} enables decoupled execution of VLM reasoning and high-frequency planning via an asynchronous parallel framework, nevertheless, APEX suffers from low absolute success rates, rendering it far from practical applicability.

Despite these advances, existing dual-system models often exhibit a relatively loose coupling between high-level semantic planning and low-level action execution, which can result in temporal misalignment and suboptimal coordination efficiency.
In aerial VLN, flight dynamics are particularly sensitive to motion continuity, and improper synchronization between reasoning and control modules may induce trajectory instability (e.g. oscillation).
In contrast, the proposed FSD-VLN introduces an explicitly coordinated fast-slow dual-system framework combined with temporally conditioned generative action modeling, aiming to balance semantic reasoning with stable and efficient low-level flight control.

\section{Methods}

\subsection{Problem Formulation}
We formulate long-horizon aerial VLN as a sequential decision process conditioned on visual observations and language instructions.
Given a natural language instruction $L$ and a sequence of visual observations $\{I_t\}_{t=1}^T$, the UAV aims to generate a sequence of flight actions $\{a_t\}_{t=1}^T$, that navigates the agent to the target location.
The objective can be expressed as modeling the conditional distribution:
\begin{align}
  p(a_{1:T} \mid  I_{1:T}, L) = p(a_1 \mid I_{1}, L) \prod_{t=2}^{T} p(a_{t} \mid a_{1:t-1} ,I_{1:t}, L).
  \label{eq:action_prob_seq}
\end{align}
In long-distance aerial navigation, action generation must satisfy two requirements: semantic consistency with global instruction guidance and temporal coherence under continuous flight dynamics.
We argue that directly modeling this distribution using a single monolithic architecture introduces structural tension between reasoning depth and control frequency.
To address this issue, we propose a fast-slow dual-system decomposition.

\begin{figure}[tb]
  \centering
  \includegraphics[width=\linewidth,page=2]{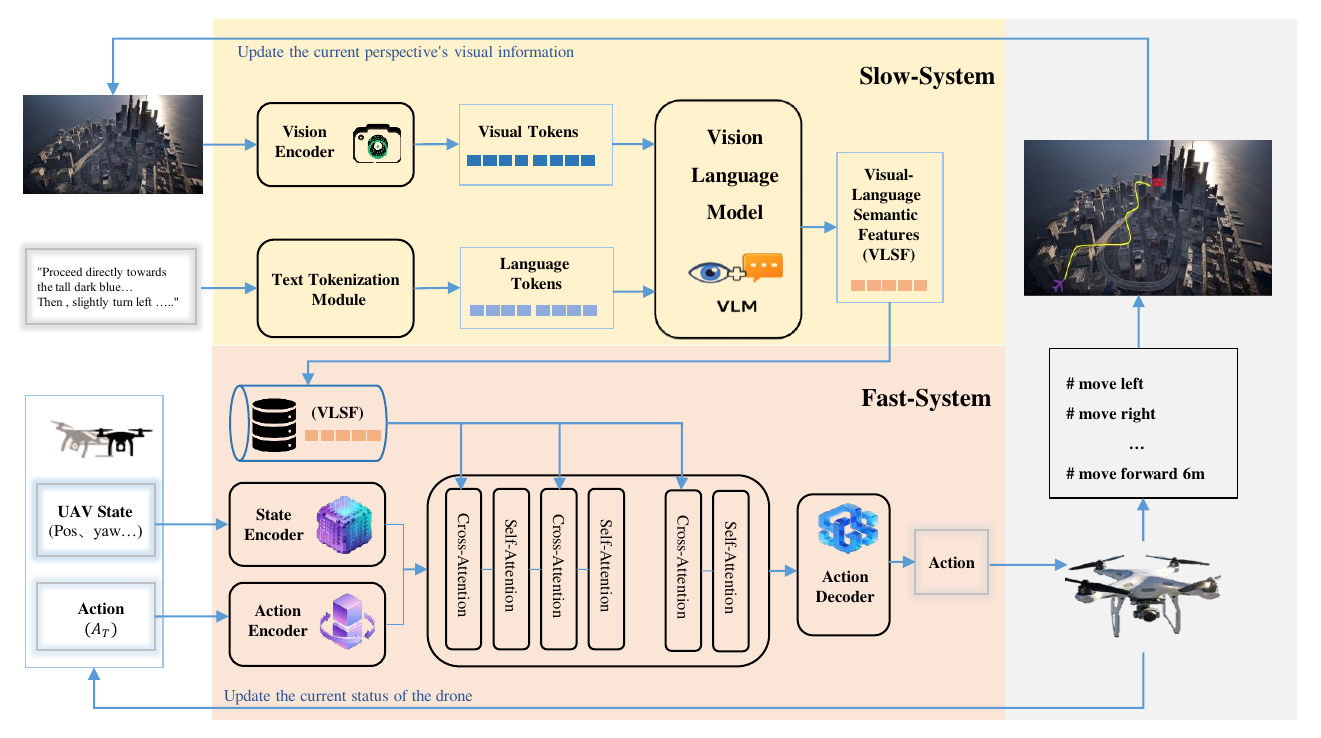}
  \caption{FSD-VLN architecture:The Slow-System processes visual and language tokens through VIM and fuses them in VLSF, producing semantic representations. The Fast-System’s DiT encodes UAV state and action priors, integrates them with VLSF features, and outputs actions via the Action Decoder, while updated visual observations feed back to the Slow-System, forming a continuous perception-action loop.
  }
\label{fig:Architecture}
\end{figure}

\subsection{Fast-Slow Dual-System Architecture}
Our FSD-VLN framework decomposes the decision process into two coordinated systems, where the Slow System captures high-level semantic priors and the Fast System generates actions through temporally conditioned modeling.

As shown in \cref{fig:Architecture}, the slow system extracts high-level multimodal semantic representations from visual observations and language instructions.
Each image frame $I_t$ is encoded by a vision encoder into visual tokens. These tokens are fused with textual embeddings and processed by a pretrained vision-language model, producing multimodal features:
\begin{align}
  z_t = f_{\text{VLM}}(I_t, L).
  \label{eq:vlm_feature}
\end{align}
The semantic feature $z_t$ acts as a global contextual prior that guides subsequent action generation.
A global cache maintains the latest semantic representation to ensure coordination between perception and control.

As show in \cref{alg:fdm}, the function of the Slow System Workflow illustrates the computation process.
Firstly, the slow system encodes the language instruction $T$ and the RGB observation $I$ to obtain the language representation $L$ and visual tokens $V$, respectively.
Then, VLM is employed to perform cross-modal fusion between linguistic semantics and visual features, producing a joint vision-language semantic representation $VL_{sf}$.
This representation is continuously written into a Vision-Language Semantic Feature buffer (VLSF). 
Whenever new visual observations arrive, the buffer is accordingly updated, thereby maintaining a stable and temporally consistent semantic representation of the environment. 
This high-level semantic prior serves as guidance for subsequent action generation.

The fast system consists of a diffusion model and three distinct multilayer perceptrons(MLPs)(State and Action Encoder, Action Decoder), building upon the diffusion transformer (DiT) architecture proposed by Peebles et al.\cite{peebles2023scalable}, we adopt a variant of the DiT module tailored to our task\cite{bjorck2025gr00t}.Specifically, the variant of DiT architecture alternates between self-attention and cross-attention layers. The self-attention layers are responsible for modeling dependencies within the agent’s state representations, enabling effective reasoning over temporal and structural correlations. In contrast, the cross-attention layers incorporate visual-language features extracted by the slow system, allowing the diffusion model to condition action generation on high-level semantic cues. The overall model is composed of multiple such attention blocks, progressively fusing the unmanned aerial vehicle (UAV) state information with visual-language representations.
At each timestep, the fast system generates an action sequence of horizon $H$:
\begin{align}
  a_{t:t+H} \sim p_0\left(a_{t:t+H} \mid z_t, s_t, a_{t-1}\right),
\label{eq:action_sampling}
\end{align}
where $s_t$ denotes the UAV state (position and yaw), $a_{t-1}$ denotes the previous action, $z_t$ is the visual-semantic prior from the slow system.
We employ a Diffusion Transformer(DiT) to model this conditional distribution.
During training, Gaussian noise is progressively added to ground-truth actions:
\begin{align}
  a_{t}^k = \sqrt{a_k} \cdot a_t + \sqrt{1 - a_k} \cdot \epsilon,\ \epsilon \sim \mathcal{N}(0, I).
\label{eq:action_aug}
\end{align}
Where $k$ denotes the denoising dimension of the DiT model.The DiT learns to denoise the corrupted action sequence, conditioned on $z_t$ state embeddings and previous action embeddings.

\begin{algorithm}[t]
\footnotesize
\caption{FSD-VLN Dual-System Workflow}
\label{alg:fdm}

\begin{algorithmic}[1]

\State \textbf{Input:}
\State \quad RGB Image $I$; Instruction $L$; State $S$; Action $A$

\State \textbf{Modules:}
\State$\begin{tabular}{p{5cm} p{5cm}}
  \quad {$VED$}: Vision Encoder & {$TTM$}: Text Tokenization Module \\
\end{tabular}$

\State$\begin{tabular}{p{5cm} p{5cm}}
  \quad {$VLM$}: Vision-Language Model & {$SED$}: State Encoder \\
\end{tabular}$

\State$\begin{tabular}{p{5cm} p{5cm}}
  \quad {$AED$}: Action Encoder & {$ADD$}: Action Decoder \\
\end{tabular}$
  

\State$\begin{tabular}{p{10cm}}
  \quad {$VLSF$}: Vision-Language Semantic Feature Buffer
\end{tabular}$

\Statex
\State \textbf{Function Slow System Workflow ($M_{SSM}$)}
\State $L_{tok} \gets \text{TTM}(L)$
\While{new RGB image $I$ arrives}
    \State $I_{tok} \gets \text{VED}(I)$
    \State $VL_{sf} \gets \text{VLM}(L_{tok}, I_{tok})$
    \State Write $VL_{sf}$ into VLSF
    \State Update $VL_{sf}^{up}$ if buffer refreshed
\EndWhile

\Statex
\State \textbf{Function Fast System Workflow ($M_{FS}$)}
\State Randomly initialize noisy actions $A$
\While{$A$ updating}
    \State $S_{seq} \gets \text{SED}(S(pos,yaw))$
    \State $A_{seq} \gets \text{AED}(A)$
    \If{VLSF updated}
        \State $d_f \gets \text{DiT}(S_{seq}, A_{seq}, VL_{sf}^{up})$
    \Else
        \State $d_f \gets \text{DiT}(S_{seq}, A_{seq}, VL_{sf})$
    \EndIf
    \State $A_{up} \gets \text{ADD}(d_f)$
\EndWhile

\Statex
\State \textbf{Function Dual-System Parallel Execution}
\State \quad $M_{SSM}(I,T)$ and $M_{FS}(S,A,VL_{sf})$ run in parallel
\State \quad UAV reaches waypoint $\Rightarrow$ update RGB image $I$

\end{algorithmic}
\end{algorithm}

As described in \cref{alg:fdm} (Function Fast System Workflow), the fast system performs temporally conditioned action generation.
The state encoder($SED$) encodes positional information ($pos,yaw$) into a state sequence representation $S_{seq}$, while the action encoder($AED$) embeds the historical action sequence into$A_{seq}$.
At each update step, if the semantic buffer has been refreshed, the most recent semantic feature $VL_{sf}^{up}$ is used; otherwise, the current cached feature $VL_{sf}$ is adopted.
These features, together with$S_{seq}$ and $A_{seq}$ are fed into a diffusion-based temporal modeling module ($DiT$) for joint generation, producing a latent representation $d_{f}$, the action decoder($ADD$) then generates the updated action $A_{up}$.
This iterative refinement process continues until convergence.

At the system level, the slow and fast systems operate in parallel. The slow system continuously incorporates new visual observations and updates the semantic buffer, while the fast system iteratively optimizes actions under semantic conditioning. 
When the UAV reaches a target waypoint, a new RGB observation is acquired, triggering another round of semantic update and action refinement. 
This design forms a closed-loop decision-making process, enabling stable and semantically consistent visual-language navigation.

For discrete action representation, we follow existing UAV-VLN works\cite{gao2025openfly,lee2024citynav, liu2023aerialvln} and define a discrete action set consisting of 8 flight primitives: Forward(3m, 6m, 9m), Turn Left(30°), Turn Right(30°), Ascend(3m), Descend(3m) and Stop. 
Although the diffusion model operates in a continuous embedding space, outputs are mapped to discrete action intervals via predefined thresholds.
This hybrid design maintains generative flexibility while preserving interpretable control commands.

\begin{figure}[tb]
  \centering
  \includegraphics[width=\linewidth,page=3]{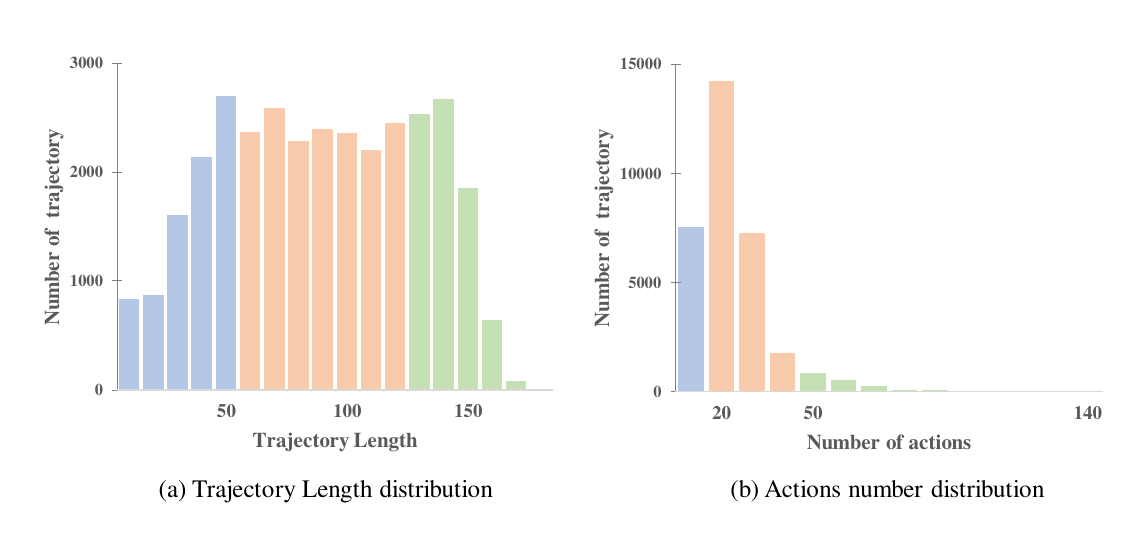}
  \caption{Statistical analysis of trajectories on the dataset.
  }
\label{fig:dataset}
\end{figure}

\subsection{Training method}
\textbf{Global Adaptive Normalization}\quad To stabilize multimodal training across heterogeneous state and action scales, we apply dataset-level statistical normalization. For each dimension $d$:
\begin{align}
  \tilde{x}_d = \frac{x_d - \mu_d}{\sigma_d},
\label{eq:feature_norm}
\end{align}
where $\mu_d$ and $\sigma_d$ denote the mean and the standard deviation respectively.
Outliers are clipped using percentile-based thresholds to mitigate long-tail effects.
During inference, predictions are rescaled to the physical control domain via inverse normalization.
This normalization ensures numerical stability across long-horizon diffusion training.
\begin{flushleft}
\textbf{Time-Weighted Optimization}\quad Long-horizon action prediction often suffers from gradient oscillation and error accumulation. To address this, we introduce a time-weighted MSE objective:
\begin{align}
  \mathcal{L}_{\text{TW-MSE}} = \frac{1}{\sum_{b,t} m_{b,t}} \sum_{b=0}^B \sum_{t=0}^T \sum_{h=0}^H w_t \cdot m_{b,t} \cdot \|\hat{\boldsymbol{a}}_{b,t,h} - \boldsymbol{a}_{b,t,h}\|^2,
\label{eq:tw_mse_loss}
\end{align}
where $B$ is batch size, $H$ is action horizon, $m_{b,t}$ is a timestep mask, and $w_t$ is the weight coefficient corresponding to the time step.
By assigning larger weights to later timesteps, the model is encouraged to maintain long-range consistency while preserving short-term accuracy.
\end{flushleft}

\section{Experiments}
\label{sec:blind}
\subsection{Evaluation Protocol}
We evaluate navigation performance using four standard metrics\cite{gao2025openfly}, namely, navigation error (NE), success rate (SR), oracle success rate (OSR), and success weighted by path length (SPL).
NE is the Euclidean distance between the final UAV position and the target location. 
SR measures the percentage of trajectories terminating within 20 meters of the target.
OSR calculates the percentage of trajectories that reach within 20 meters of the target at any timestep.
SPL is a metric that jointly measures success and path efficiency by weighting SR with the ratio of the ground-truth path length.
These metrics collectively assess accuracy, robustness, and trajectory quality in long-horizon aerial navigation. 

\begin{table}[tb]
  \caption{Comparison of navigation performance among different baseline models. We bold the best results and underline the second best results.
  }
  \label{tab:t1}
  \centering
  \tabcolsep=12pt
  \begin{tabular}{@{}ccccc@{}}
    \toprule
    \multicolumn{4}{c}{Unseen} \\
    \cmidrule{2-5}
     &  SR$\uparrow$  & OSR$\uparrow$ & SPL$\uparrow$ & NE$\downarrow$  \\
    \midrule
    Random  & 0.1\% & 0.1\% & 0\%   & 301 \\
    CMA\cite{krantz2020beyond} & 0.7\% & 6.6\% & 0.7\% & 268 \\
    Seq2Seq\cite{krantz2020beyond} & 0.1\% & 3.2\% & 0.6\% & 291 \\
    Navid\cite{zhang2024navid} & 2.6\% & 16.9\% & 2.0\% & 200 \\
    AeriaVLM\cite{liu2023aerialvln} & 2.3\% & 19.8\% & 2.2\% & 227 \\
    CityNavAgent\cite{zhang2025citynavagent} & \underline{11.7\%} & \bf{35.2}\% & 5.0\% & \bf{60} \\
    OpenFly\cite{gao2025openfly}\textsuperscript{*} & 5.1\% & 27.3\% & 3.5\% & 198 \\
    \textbf{FSD-VLN (ours)} & \bf{13.6}\% & 28.4\% & \bf{10.7}\% & \underline{78} \\
    \toprule
    \multicolumn{4}{c}{Seen} \\
    \cmidrule{2-5}
     &  SR$\uparrow$  & OSR$\uparrow$ & SPL$\uparrow$ & NE$\downarrow$  \\
    \midrule
    Random   & 0.7\% & 0.8\%    & 0\% & 242\\
    CMA\cite{krantz2020beyond}  & 5.7\% & 30.0\% & 5.1\% & 144\\
    Seq2Seq\cite{krantz2020beyond}  & 4.0\% & 16.8\% & 3.1\% & 194\\
    Navid\cite{zhang2024navid}  & 9.9\% & 24.3\% & 3.9\% & 142\\
    AeriaVLM\cite{liu2023aerialvln}  & 6.6\% & 40.8\% & 7.0\% & 103\\
    CityNavAgent\cite{zhang2025citynavagent}  & 13.9\% & 30.2\% & 10.2\% & \bf{61}\\
    OpenFly\cite{gao2025openfly}\textsuperscript{*}  &\underline{18.5\%} & \bf{50.9}\% & \underline{12.2\%} & 115 \\
    \textbf{FSD-VLN (ours)}  & \bf{26.7}\% & \underline{43.3\%} & \bf{22.8}\% & \underline{76}\\
  \bottomrule
  \end{tabular}
  \vspace{0.1em}
  \parbox{\linewidth}{\footnotesize \noindent \centering $^*$ Self-reproduced results using the official code on our test set.}
\end{table}

\subsection{Datasets and Implementation Details}
We evaluate our approach in large-scale simulated urban environments constructed from AirVLN-S and OpenFly. To ensure generality, we randomly sampled datasets from four virtual city scenarios and additionally incorporated the Guangzhou urban rendering scene from OpenFly\cite{gao2025openfly}, resulting in over 30,000 navigation trajectories spanning diverse urban settings, including downtown, industrial areas, parks, and villages. Trajectory statistics (\cref{fig:dataset}) show lengths concentrated in 50–150 meters (Figure \cref{fig:dataset}(a)) and action counts in the 20–50 range after merging consecutive ``move forward 3 m'' actions into single “move forward 9 m” actions (\cref{fig:dataset}(b)), highlighting realistic navigation distances and sequence complexity. All data were collected using a high-fidelity Unreal Engine simulator.

Models are initialized with the GR00T N1 pre-trained backbone. During fine-tuning, only the DiT decision module, state encoder, and action decoder are updated, while the visual-language encoder is frozen to preserve general multimodal representations.The slow system employs 16 DiT blocks, with Output-dim and Input-embedding-dim set to 1024 and 1536, respectively, and Hidden-size set to 1024. We optimize with AdamW (weight decay $10^{-5}$) and LoRA (alpha 16) for 200K steps on 4 RTX 4090 GPUs, requiring approximately 320 GPU hours.

\begin{figure}[tb]
  \centering
  \includegraphics[width=\linewidth,page=4]{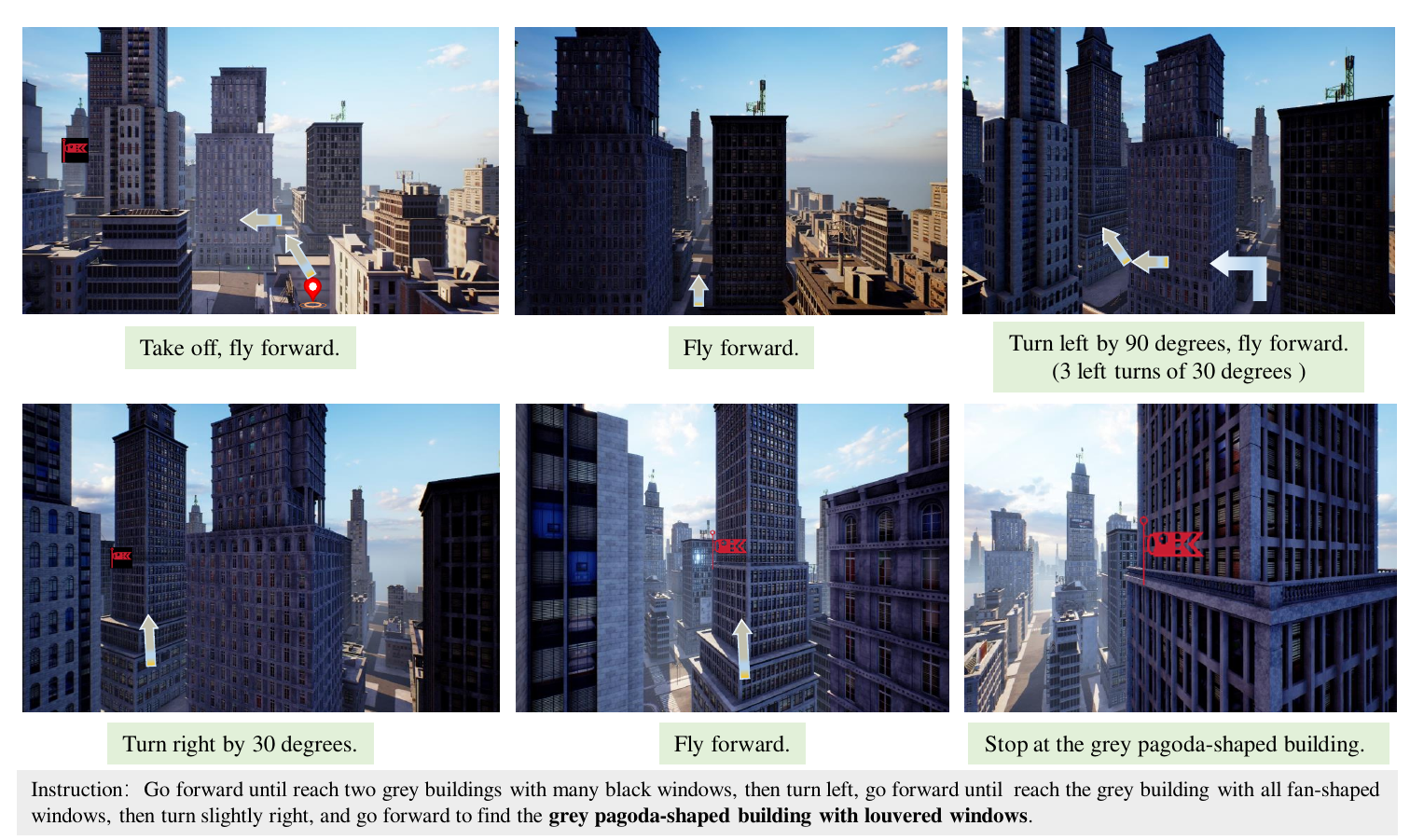}
  \caption{Example aerial navigation:The agent ascends to the initial altitude, adjusts its pitch, and parses the instruction. It flies forward sequentially, reaching the location in the rightmost image of the first row to identify the landmark “two grey buildings with many black windows” and executes a $90^\circ$ left turn. Continuing along the updated heading, it recognizes the gray building with fan-shaped windows in the leftmost image of the second row and applies a $30^\circ$ right yaw adjustment. The agent then proceeds to the target landmark, “the grey pagoda-shaped building with louvered windows,” where it stops, successfully completing the navigation task.
  }
\label{fig:Qualitative}
\end{figure}

\subsection{Main Results}
The comparison with representative baselines including Random, Seq2Seq\cite{krantz2020beyond}, Naid\cite{zhang2024navid}, AerialVLN\cite{liu2023aerialvln}, and OpenFly\cite{gao2025openfly} are presented in \cref{tab:t1}.
In the ``Random'' method, the agent randomly samples and executes actions from the action set, terminating upon selecting the stop action or reaching the maximum step limit.
FSD-VLN consistently outperforms prior methods in both seen and unseen environments.
Notably, in unseen scenarios, FSD-VLN significantly outperforms OpenFly, improving SR from 5.1\% to 13.6\% and SPL from 3.5\% to 10.7\%, while reducing NE from 198 to 78, indicating more stable and globally consistent trajectory planning.
The substantial reduction in navigation error indicates improved global trajectory alignment rather than marginal success improvements.
In seen environments, FSD-VLN maintains superior SR (26.7\%) and SPL (22.8\%), demonstrating that the proposed framework does not sacrifice efficiency for robustness.

Unlike reactive baselines that generate actions solely based on instantaneous vision-language observations, FSD-VLN formulates action prediction as a temporally conditioned generative process.
Through structured diffusion modeling, current decisions are explicitly grounded in historical action trajectories.
This mechanism alleviates the myopic perception bias caused by frame-wise reasoning and suppresses oscillatory behaviors induced by over-reactivity to transient visual signals.
More importantly, by maintaining temporal coherence in the latent action space, the model effectively mitigates the compounding error that commonly arises in long-horizon navigation.
The consistent reduction in NE across both seen and unseen environments demonstrates that the fast-slow decomposition improves global direction consistency and trajectory stability, rather than merely increasing the terminal success rate.

To evaluate real-time capability, we compare FSD-VLN with OpenFly under comparable model scales. As shown in \cref{tab:t2}, FSD-VLN reduces per-action inference latency from 402 ms to 176 ms, demonstrating substantially improved decision efficiency.
Beyond single-step latency, we further analyze end-to-end navigation time on 214 randomly sampled trajectories. As presented in \cref{tab:t3}, the total execution time decreases from 307.6 s to 144.7 s, corresponding to a 53\% reduction, while the average number of executed actions is reduced by approximately 10\%.
To control for potential performance bias, we additionally evaluate 10 trajectories that are successfully accomplished by both methods.
Even under identical success conditions, FSD-VLN achieves a 56\% reduction in execution time.
These results indicate that the efficiency improvement arises not only from reduced per-step inference overhead but also from enhanced trajectory-level consistency, which mitigates redundant corrective maneuvers and stabilizes long-horizon decision dynamics.

\begin{table}[tb]
  \caption{Test results of action generation time consumption by the model.
  }
  \label{tab:t2}
  \centering
  \tabcolsep=12pt
  \begin{tabular}{@{}c|cc@{}}
    \toprule
    & OpenFly & Ours \\
    \midrule
    Single Action Generation(ms)  & 402 & 176 \\
    Data Preparation and Action Generation(ms) & 576 & 387 \\
  \bottomrule
  \end{tabular}
\end{table}

\cref{fig:Qualitative} presents qualitative examples of successful navigation flights achieved by the proposed Fast-Slow Dual-System Modeling(FSD-VLM) framework using landmark-based visual map information. The examples show how the aerial agent leverages intermediate visual landmarks—such as “two grey buildings with many black windows” and the gray building with fan-shaped windows—to sequentially adjust its heading, progressively localizing the target.The examples validate the effectiveness of FSD‑VLM in integrating visual semantic information with low-level flight control.

\begin{table}[tb]
  \caption{Quantitative Evaluation of Action Generation Efficiency Between OpenFly and our proposed FSD-VLN method.
  }
  \label{tab:t3}
  \centering
  \tabcolsep=12pt
  \begin{tabular}{@{}c|cc|cc@{}}
    \toprule
    & \multicolumn{2}{c}{OpenFly} & \multicolumn{2}{c}{Ours} \\
    & Total steps & Time(s) & Total steps & Time(s)  \\
    \midrule
    Test set (214)  & 4992 & 307.62& 4468 & 144.72\\
    \makecell{Successfully completed \\ identical routes(10)}& -- & 52.71 & -- & 22.78\\
  \bottomrule
  \end{tabular}
\end{table}

In this task, the aerial agent first ascends to the initial altitude and adjusts its flight pitch. 
It then parses the given instruction and, conditioned on the current visual observation, sequentially executes multiple forward flight actions. 
Upon reaching the geographic location shown in the rightmost image of the first row, the agent identifies the landmark described as “two grey buildings with many black windows” and subsequently performs a $90^\circ$ left turn.

The agent continues flying straight along the updated heading. 
When arriving at the position shown in the leftmost image of the second row, it successfully recognizes the gray building with fan-shaped windows described in the instruction and executes a $30^\circ$ right yaw adjustment. After orientation correction, the agent proceeds forward until reaching the target landmark described as “the grey pagoda-shaped building with louvered windows,” where it stops. 
The navigation task is thus successfully completed.

\subsection{Ablation Study}
\begin{figure}[tb]
  \centering
  \includegraphics[width=\linewidth,page=5]{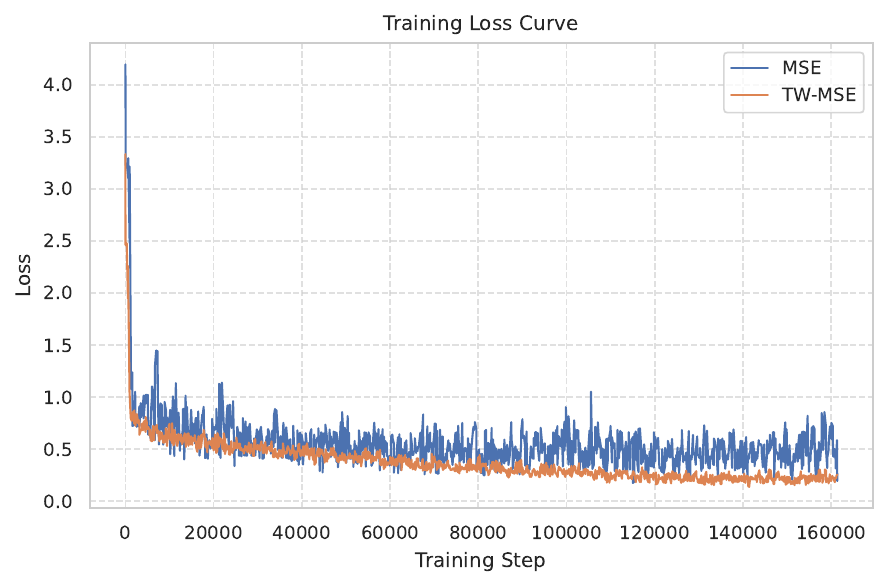}
  \caption{Training loss curves of the standard MSE (a) or TW-MSE (b) loss functions.
  }
  \label{fig:loss}
\end{figure}

\textbf{Effect of Time-Weighted Optimization}\quad To stabilize long-horizon action prediction, we introduce a Time-Weighted MSE (TW-MSE) loss during fine-tuning, which assigns temporally adaptive weights to prediction errors across different time steps.
Compared with standard MSE, TW-MSE explicitly models the varying importance of actions along the temporal dimension.
The \cref{fig:loss} presents the training loss curves under MSE and TW-MSE. TW-MSE exhibits smoother convergence behavior, reduced oscillation amplitude, and a lower final loss.
The improvement can be attributed to its temporal reweighting mechanism, which suppresses gradient interference from distant future predictions and alleviates error amplification in long sequences.
From an optimization perspective, TW-MSE introduces an implicit temporal regularization that stabilizes gradient propagation across time steps, thereby mitigating common issues of gradient fluctuation and error accumulation in sequential decision modeling.
Beyond optimization stability, TW-MSE also improves final navigation performance, indicating that temporally aware supervision enhances the model's ability to balance short-term action precision and long-term trajectory planning.
These results validate the effectiveness of time-aware weighting strategies for long-horizon UAV visual-language navigation.

\textbf{Effect of Action Horizon}\quad The action-horizon $H$ parameter controls the number of future actions generated per inference step, reflecting the temporal planning span of the model. 
Increasing action-horizon enables longer-horizon foresight and potentially reduces inference frequency, which may lower decision latency.
However, in dynamic environments with rapidly changing visual observations and UAV states, long-span prediction introduces greater distribution shift and compounding error risks.
To investigate this trade-off, we evaluate $H \in \{1, 2, 4\}$ in both seen and unseen environments over 154 trajectories of varying complexity.
The results are summarized in \cref{tab:t4}. Performance consistently degrades as action-horizon increases, with $H = 1$ achieving the best SR and SPL as well as the lowest NE.
Although longer horizons theoretically provide stronger planning capability, they amplify prediction bias due to perceptual latency and environmental non-stationarity.
As the generated action sequence extends further into the future, the discrepancy between predicted and actual state distributions accumulates, leading to reduced trajectory stability.
This finding highlights a fundamental trade-off between long-horizon planning and real-time perceptual responsiveness in UAV visual-language navigation.
While multi-step prediction can amortize inference cost, excessive temporal extrapolation undermines robustness in highly dynamic settings.
Therefore, maintaining short-horizon reactivity while leveraging structured temporal modeling emerges as a more stable strategy for long-distance aerial navigation.

\begin{table}[]
  \caption{Impact of Action Horizon on Navigation Performance for FSD-VLN.
  }
  \label{tab:t4}
  \centering
  \tabcolsep=12pt
  \begin{tabular}{@{}cc|ccccc@{}}
    \toprule
    & $H$ & SR & OSR & SPL & NE \\
    \midrule
    & 1 & 20.13\% & 44.16\% & 18.06\% & 88 &\\
    & 2 & 16.88\% & 39.61\% & 14.87\% & 100 &\\
    & 4 & 15.58\% & 44.80\% & 13.50\% & 90 &\\
  \bottomrule
  \end{tabular}
\end{table}

\section{Conclusion}
In this work, we address key challenges in long-distance low-altitude UAV visual-language navigation, including high decision latency, limited action continuity, and difficulties in long-horizon modeling.
We propose FSD-VLN, an efficient fast-slow dual-system framework that explicitly coordinates high-level semantic reasoning with low-level action generation.
By integrating multimodal perception, temporally structured decision modeling, and real-time control, the proposed framework provides a unified solution for autonomous navigation in complex open environments.

In addition, we introduce a global adaptive preprocessing strategy to mitigate scale and distribution discrepancies across multimodal inputs, together with a time-aware weighted optimization objective that stabilizes long-sequence training.
Extensive experiments in both seen and unseen simulated environments demonstrate that FSD-VLN consistently improves navigation accuracy, trajectory quality, and decision efficiency while substantially reducing inference latency and overall execution time.
These results indicate that structured fast-slow decomposition offers an effective paradigm to balance global planning consistency and real-time responsiveness in UAV-VLN.

Despite these improvements, several limitations remain. Performance may degrade under highly dynamic environmental changes, where perception delay and distribution shift pose challenges for long-horizon prediction.
Future work will explore scaling the framework with larger multimodal foundation models and broader datasets to enhance cross-scene generalization.
We also aim to extend FSD-VLN to multi-UAV coordination and real-world deployment, further bridging the gap between simulated benchmarks and practical aerial navigation systems.

\bibliographystyle{splncs04}
\bibliography{arxiv}

%
%

\end{document}